# GSSF: A Generative Sequence Similarity Function based on a Seq2Seq model for clustering online handwritten mathematical answers

Huy Quang Ung [0000-0001-9238-8601], Cuong Tuan Nguyen [0000-0003-2556-9191], Hung Tuan Nguyen [0000-0003-4751-1302], and Masaki Nakagawa [0000-0001-7872-156X]

Tokyo University of Agriculture and Technology, Tokyo, Japan

ungquanghuy93@gmail.com
fx4102@go.tuat.ac.jp
ntuanhung@gmail.com
nakagawa@cc.tuat.ac.jp

**Abstract.** Toward a computer-assisted marking for descriptive math questions, this paper presents clustering of online handwritten mathematical expressions (OnHMEs) to help human markers to mark them efficiently and reliably. We propose a generative sequence similarity function for computing a similarity score of two OnHMEs based on a sequence-to-sequence OnHME recognizer. Each OnHME is represented by a similarity-based representation (SbR) vector. The SbR matrix is inputted to the $k$-means algorithm for clustering OnHMEs. Experiments are conducted on an answer dataset (Dset_Mix) of 200 OnHMEs mixed of real patterns and synthesized patterns for each of 10 questions and a real online handwritten mathematical answer dataset of 122 student answers at most for each of 15 questions (NIER_CBT). The best clustering results achieved around 0.916 and 0.915 for purity, and around 0.556 and 0.702 for the marking cost on Dset_Mix and NIER_CBT, respectively. Our method currently outperforms the previous methods for clustering HMEs.

## 1    Introduction

In 2020, the widespread of the SARS-CoV-2 (COVID-19) has a strong impact on education over the world, which caused almost schools and universities to be temporally closed. Many educational organizers resume the learners' studies via online platforms in response to significant demand. The adoption of online learning might continue persisting in post-pandemic, and the shift would impact the worldwide education market. In this context, the self-learning and e-testing applications would be necessary options for students in the near future.

Nowadays, touch-based and pen-based devices are widely used as learning media. Learners use them to read textbooks, study online courses and do exercise. Moreover, those devices are suitable for learners to quickly input mathematical expressions (MEs), which seems better than using common editors such as Microsoft Equation Editor, MathType, and LaTeX.

Over the last two decades, the study of online handwritten mathematical expressions (OnHMEs) recognition has been actively carried out due to demands for its application on tablets. Competitions on recognizing OnHMEs have been ongoing under the series of CROHME with better recognition performance [1].In this context, many e-learning interfaces based on pen-based devices have been studied and adopted in practical applications [2][3][4]. If learners can verify and confirm the recognition results of their answers, we can incorporate the handwritten mathematical expression (HME) recognition into self-learning and e-testing applications. Although learners have to do additional work, they can receive immediate feedback.

HME recognition can also be used for marking. Automatically marking answers by comparing the recognition result of an HME answer with the correct answer is one solution. However, there remain the following problems as mentioned in [5]. Firstly, it is complex to mark partially correct answers. Secondly, there may be several correct answers as well as some different but equivalent notations for an answer (e.g., "a + b" and "b + a"). Therefore, it is not practical to pre-define all possible cases. Thirdly, it requires examiners or examinees to confirm the automatic marking since the recognition results are not entirely perfect. However, large-scale examinations (e.g., national-wide qualification examinations) do not often provide opportunities for the examinees to confirm the marking, so examiners or someone else must confirm the marking.

Computer-assisted marking is an alternative approach for marking HME answers. Instead of recognizing, answers are clustered into groups of similar answers then marked by human markers. Ideally, when the answers in each group are the same, it takes only one action to mark all the answers for each group. It reduces the huge amount of marking efforts and reduces the marking errors. Since examinees make the final marking, their anxieties will also decrease.

Although clustering HMEs is promising for computer-assisted marking, the approach encounters two main challenges. Firstly, it is challenging to extract the features that represent the two-dimensional structure of HMEs. The bag-of-features approach [5, 6] combined low-level features such as directional features and high-level features such as bag-of-symbols, bag-of-relations and bag-of-position. These features, however, rather capture the local features instead of the global structure of HMEs. Secondly, we need a suitable similarity/dissimilarity measurement between the feature representations of two HMEs, in which the previous research utilized Euclidean distance [5, 6] or cosine distance [7]. However, Euclidean distance and $p$-norm seem ineffective in a high-dimensional space because of the concentration phenomenon, in which all pairwise distances among data-points seem to be very similar [8]. These approaches are also limited to measure the fixed-size features.

This paper proposes a method that utilizes a generative sequence similarity function (GSSF) and a data-driven representation for each OnHME. GSSF is formed by high-level sequential features, probability terms of the output sequence generated from a sequence-to-sequence (Seq2Seq) OnHME recognizer. The sequential features are dynamic, and they could represent the global structure of OnHME. Each OnHME is then represented by a vector of similarity scores with other OnHMEs, namely similarity-based representation (SbR). SbR allows controlling the dimensionality of the feature space to reduce the influence of the concentration phenomenon. Finally, we input the SbR matrix into a clustering algorithm such as $k$-means to obtain the clusters of OnHMEs.

The rest of the paper is organized as follows. Section 2 briefly presents related research. Section 3 describes our method in detail. Section 4 focuses on problems related to the cost of clustering-based marking. Section 5 presents our experiments for evaluating the proposed method. Finally, section 6 concludes our work and discusses future works.

## 2 Related works

This section presents previous studies of computer-assisted marking and Seq2Seq-based OnHME recognition.

### 2.1 Computer-assisted marking

There are several past studies on clustering offline (bitmap image) HMEs (OfHMEs). Khuong et al. [5] also combined low-level features and high-level features (bag-of-symbols, bag-of-relations, and bag-of-positions) to represent each OfHME. However, those high-level features are formed by recognizing offline isolated symbols from connected components in an OfHME along with pre-defined heuristic rules. Hence, there are problems related to segmentation and determination of spatial relationships. Nguyen et al. [7] presented features based on hierarchical spatial classification using a convolutional neural network (CNN). Their CNN model is trained to localize and classify symbols in each OfHME using weakly supervised learning. Then, spatial pooling is applied to extract those features from the class activation maps of the CNN model. The authors also proposed a pairwise dissimilarity representation for each OfHME by computing the cosine distance between two OfHMEs.

For essay assessment [9–11] and handwritten essay scoring [12], extensive research has been conducted. Basu et al. also proposed a method for clustering English short answers [13]. They formed a similarity metric by computing a distance between two answers using logistic regression. Then, they utilized a modified k-Medoids and a latent Dirichlet allocation algorithm for forming clusters of answers. As an extended work of [13], Brooks et al. designed a cluster-based interface and demonstrated its effectiveness [14].

There are deep neural network-based methods for clustering sequential data such as OnHMEs. Several methods aim to embed sequential data into feature vectors based on the reconstruction loss and the clustering loss [24-25]. Another approach is to compute the pairwise similarity/dissimilarity instead of embedded features [26]. However, those methods without information about symbols and relations encounter difficulty for clustering OnHMEs since there are infinite combinations of symbols and relations to form mathematical expression (MEs). Nguyen et al. [7] showed that met-ric learning methods do not work well compared to CNN-based spatial classification features for clustering offline HMEs.

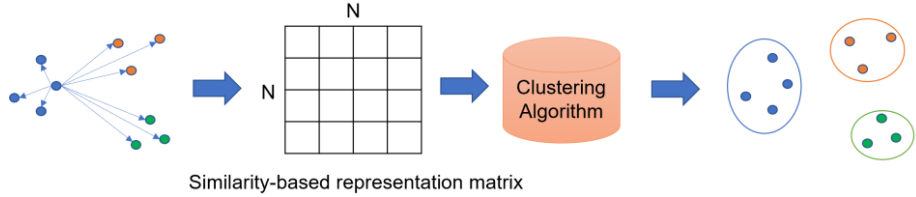

**Fig. 1.** Clustering process of our proposed method

### 2.2 Seq2seq OnHME recognizer

Seq2Seq models for recognizing OnHMEs are deep learning models that directly convert an input sequence into an output sequence. A Seq2Seq model consists of two main components, an encoder and a decoder. The encoder, an LSTM or a BLSTM network, accepts a time series input of arbitrary length and encodes information from the input into a hidden state vector. The decoder, commonly an LSTM network, generates a sequence corresponding to the input sequence.

Zhang et al. [15] proposed a track, attend and parse (TAP) architecture, which parses an OnHME into a LaTeX sequence by tracking a sequence of input points. The encoder or the tracker stacks several layers of bidirectional Gated Recurrent Units (GRUs) to get the high-level representation. The decoder or the parser is composed of unidirectional GRUs combined with a hybrid attention mechanism and a GRU-based language model. The hybrid attention consists of two attentions: coverage based spatial attention and temporal attention. Hong et al. [16] improved TAP by adding residual connections in the encoder to strengthen the feature extraction and jointly using a transition probability matrix in the decoder to learn the long-term dependencies of mathematical symbols.

## 3 Proposed method

Cluster analysis is useful for exploratory analysis by partitioning unlabeled samples into meaningful groups. For this problem, we traditionally extract useful features from each sample and pass them to a clustering algorithm such as $k$-means to partition them into groups. In this paper, we utilize a type of data-driven representation for each sample. We represent each sample by pairwise similarities between it and other samples. Then, we form a similarity-based representation (SbR) matrix. The SbR matrix is inputted to a clustering algorithm to obtain clusters of OnHMEs. The overall process of our proposed method is shown in Fig. 1. This section presents our proposed similarity function (SF), then describes the SbR.

### 3.1 Generative sequence similarity function

Our proposed SF gives a similarity score between two OnHMEs based on a Seq2Seq OnHME recognizer. We expect that the similarity score of two OnHMEs containing the same ME is significantly higher than those with different MEs.

A standard Seq2Seq model, as shown in Fig. 2, consists of two parts: an encoder that receives an input sequence to represent high-level features and a decoder that sequentially generates an output sequence from the encoded features and the previous prediction. Given two input OnHMEs denoted as $S_1$ and $S_2$, the recognizer generates LaTeX sequences of $\{y_i^1\}_{i=\overline{1,N}}$ and $\{y_j^2\}_{j=\overline{1,M}}$, where $y_i^1$ and $y_j^2$ are symbol classes in the vocabulary, and $N$ and $M$ are the lengths of the output sequences.

A simple idea to form the similarity score of two OnHMEs is to calculate the edit distance of the two output sequences $\{y_i^1\}_i$ and $\{y_j^2\}_j$. However, this method might not be effective since the edit distance only utilizes the differences in terms of recognized symbol classes, but the probabilities of recognized symbols seem more important than the symbol classes. Our proposed SF utilizes terms of probabilities of recognized symbol classes instead of the symbol classes.

Another difficulty of directly comparing two output sequences or generated probabilities is that their lengths are variant. The proposed SF uses the symbol predictions of an OnHME to input into the decoder of another OnHME for computing the terms of probabilities. Those probabilities are formed on the output sequence of one of those two OnHMEs. Hence, the SF is not influenced by the size-variant problem.

Our SF consists of two main components: the similarity score of $S_1$ compared to $S_2$ and the one of $S_2$ compared to $S_1$ denoted as $F(S_1|S_2)$ and $F(S_2|S_1)$, respectively. We firstly define $F(S_1|S_2)$ as follows:

$$F(S_1|S_2) = \sum_{i=1}^{N} \left( \log\left(P(y_i^1|S_2, y_{i-1}^1)\right) - \log\left(P(y_i^1|S_1, y_{i-1}^1)\right) \right) \quad (1)$$

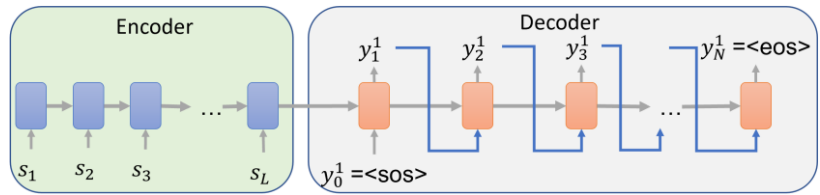

**Fig. 2.** A standard Seq2Seq model

where $P(x|y,z)$ is the probability of $x$ given y and z. An illustration of computing $F(S_1|S_2)$ is shown in Fig. 3. The predicted symbol $y_i^1$ at the $i$-th time step is inputted to the $(i+1)$-th time step of the $S_2$ decoder for computing the probability $P(y_i^1|S_2, y_{i-1}^1)$. Similarly, we define $F(S_2|S_1)$ as follows:

$$F(S_2|S_1) = \sum_{j=1}^{M} \left( \log\left( P(y_j^2|S_1, y_{j-1}^2) \right) - \log\left( P(y_j^2|S_2, y_{j-1}^2) \right) \right) \qquad (2)$$

The $F$ function is not appropriate for the clustering algorithms such as $k$-means because it is asymmetrical. Thus, we compute an average of $F(S_1|S_2)$ and $F(S_2|S_1)$ that is symmetrical measurement. We name it as the Generative Sequence Similarity Function (GSSF), which is computed as follows:

$$GSSF(S_1, S_2) = \frac{F(S_1|S_2) + F(S_2|S_1)}{2} \qquad (3)$$

Assume that the Seq2Seq recognizer are well recognized $S_1$ and $S_2$. GSSF has some properties as follows:

- $GSSF(S_1, S_1)$ equals to zero if and only if $F(S_1|S_1)$ equals to zero.
- $GSSF(S_1, S_2)$ is approximately zero if $S_1$ and $S_2$ denote the same ME. In this case, $F(S_1|S_2)$ and $F(S_2|S_1)$ are both around zero.
- $GSSF(S_1, S_2)$ is negative if $S_1$ and $S_2$ denote two different MEs. In this case, both $F(S_1|S_2)$ and $F(S_2|S_1)$ are much lower than zero.
- GSSF is a symmetric function.

### 3.2 Similarity-based representation

Given $N$ OnHMEs $\{X_1, X_2, \dots, X_N\}$, SbR of $X_i$ is formed by a pre-defined pairwise SF:

$$SbR(X_i) = [SF(X_i, X_1), \dots, SF(X_i, X_i), \dots, SF(X_i, X_N)] \qquad (4)$$

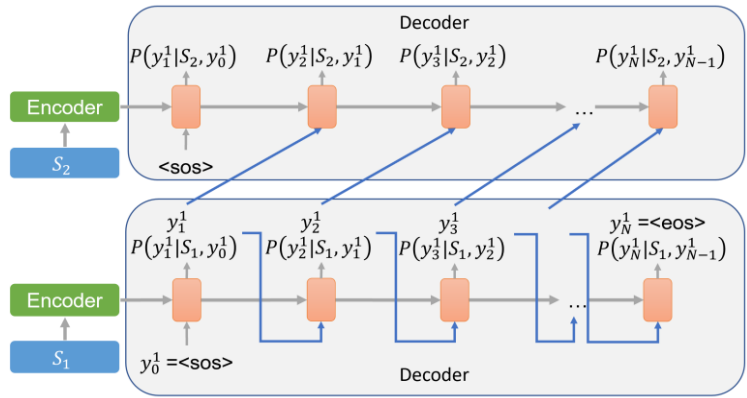

**Fig. 3.** Illustration of computing $F(S_1|S_2)$.

The values of $SbR(X_i)$ is normalized into $[0, 1]$ before inputting to the clustering algorithm.

## 4    Clustering-based measurement

In clustering-based marking systems, a human marker marks the major set of answers for each cluster collectively and selects the minor ones for manual marking separately. Hence, the cost of the marking process depends on how many samples belong to the major set and how few answers in the minor sets are included in each cluster. For this reason, we measure purity to evaluate the performance of the clustering task as shown in Eq. (5):

$$Purity(G, C) = \frac{1}{H} \sum_{k=1}^{K} \max_{1 \le i \le J} |g_k \cap c_i| \qquad (5)$$

where $H$ is the number of samples, $J$ is the number of categories (the right number of clusters), $C = \{c_1, c_2, \ldots, c_J\}$ is a set of categories, $K$ is the number of clusters, and $G = \{g_1, g_2, \ldots, g_K\}$ is a set of obtained clusters.

However, high purity is easy to achieve when the number of clusters is large. For example, when $K$ is equal to $H$, we obtain a perfect purity of 1.0. Hence, we set the number of clusters as the number of categories to evaluate in our experiments.

The purity alone does not show the quality of clustering in the clustering-based marking systems. We employ a cost function presented in Khuong et al. [5], reflecting a scenario of verifying and marking answers in the clustering-based marking systems. For each cluster, the verifying task is to find a major set of answers by filtering minor answers, while the marking task is to compare the major set and minor answers with the correct answer or the partially correct answers. The marking cost ($MC$) is composed of the verifying time $C_{ver}$ and the marking time $C_{mark}$ as shown in Eq. (6):

$$f(G, C) = \sum_{i=1}^{K} cost(g_i, C) = \sum_{i=1}^{K} \left( C_{ver}(g_i, C) + C_{mark}(g_i, C) \right)$$

$$= \sum_{i=1}^{K} (|g_i| \times \alpha T + (1 + |g_i| - |M_i|) \times T) \qquad (6)$$

where $T$ is the time unit to mark an answer. There exists a real number $\alpha$ ($0 < \alpha \le 1$) so that the verifying cost of an answer is $\alpha T$. $C_{ver}(g_i, C) = |g_i| \times \alpha T$ is the cost of verifying all answers in the cluster $g_i$. $C_{mark}(g_i, C) = (1 + |g_i| - |M_i|) \times T$ is the cost of marking the major set of answers $M_i$ and all minor answers in the cluster $g_i$. For simplicity, we assume $\alpha = 1$, implying that the verification time is the same as the marking time $T$. We normalize Eq. (6) into $[0, 1]$, and obtain Eq. (7) as follows:

$$MC(G, C) = \frac{f(G, C)}{2NT} = \frac{K}{2H} + \left(1 - \frac{1}{2} Purity(G, C)\right) \qquad (7)$$

$MC$ equals 1 in the worst case if the number of clusters equals the number of answers. It implies that $MC$ approaches the cost of marking all answers one-by-one.

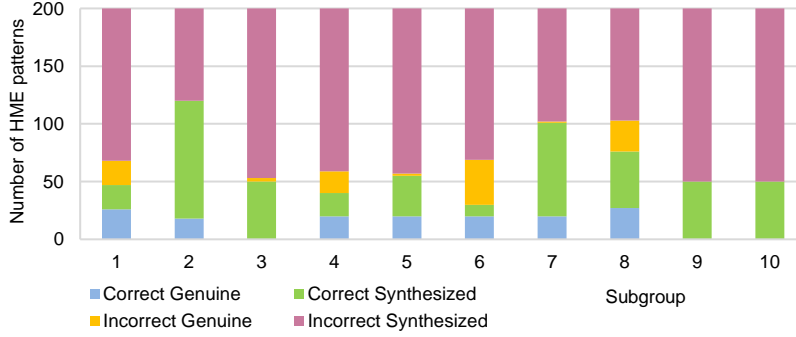

**Fig. 4.** Details of each subgroup in Dset_Mix.

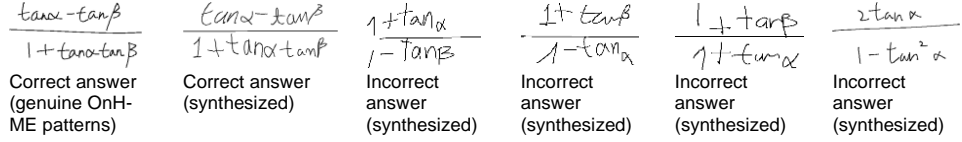

| $\frac{\tan\alpha - \tan\beta}{1 + \tan\alpha\tan\beta}$ | $\frac{\tan\alpha - \tan\beta}{1 + \tan\alpha\tan\beta}$ | $\frac{1 + \tan\alpha}{1 - \tan\beta}$ | $\frac{1 + \tan\beta}{1 - \tan\alpha}$ | $\frac{1 + \tan\beta}{1 + \tan\alpha}$ | $\frac{2\tan\alpha}{1 - \tan^2\alpha}$ |
|---|---|---|---|---|---|
| Correct answer (genuine OnHME patterns) | Correct answer (synthesized) | Incorrect answer (synthesized) | Incorrect answer (synthesized) | Incorrect answer (synthesized) | Incorrect answer (synthesized) |

**Fig. 5.** Samples in subgroup 8 of Dset_Mix.

## 5 Experiments

This section presents the evaluation of our proposed SF on two answer datasets, i.e., Dset_Mix and NIER_CBT, by using the TAP recognizer proposed in [15] without the language model. We name this modified recognizer as MTAP.

### 5.1 Datasets

We use two datasets of OnHMEs to evaluate our proposed method. The first dataset, named Dset_Mix [17], consists of mixed patterns of real (genuine) OnHMEs from CROHME 2016 and synthesized patterns. The synthesized patterns are generated according to LaTeX sequences and isolated handwritten symbol patterns from CROHME 2016. Dset_Mix consists of ten subgroups, corresponding to ten questions. Each subgroup consists of 200 OnHMEs, a mixture of genuine patterns and synthesized patterns. The synthesized OnHMEs were generated according to the method proposed in [18]. The sample size for each question is set according to the number of students in each grade of common schools. The details of each subgroup are shown in Fig. 4. Each subgroup contains a few correct answers and several incorrect answers. Fig. 5 shows some samples in subgroup 8 of Dset_Mix.

The second dataset, named NIER_CBT, is a real answer dataset collected by a collaboration with National Institute for Educational Policy Research (NIER) in Tokyo, Japan. NIER carried out math tests for 256 participants, consisting of 249 students of grade 11 and 7 students of grade 7 at nine high schools. The participants answered a set of 5 questions within 50 minutes, then wrote their results on iPad by using an Apple pen and a developed tool with OnHMEs captured. The details of the collection are presented in [19]. There are three sets of questions to obtain 934 answers for 15 questions. Since our OnHME recognizer was trained for 101 common math symbols that appeared in the dataset of CROHME [1], we removed 15 OnHMEs that contain out-of-vocab symbols. The number of correct/incorrect answer categories and the number of correct/incorrect answer patterns in NIER_CBT are shown in Fig. 6.

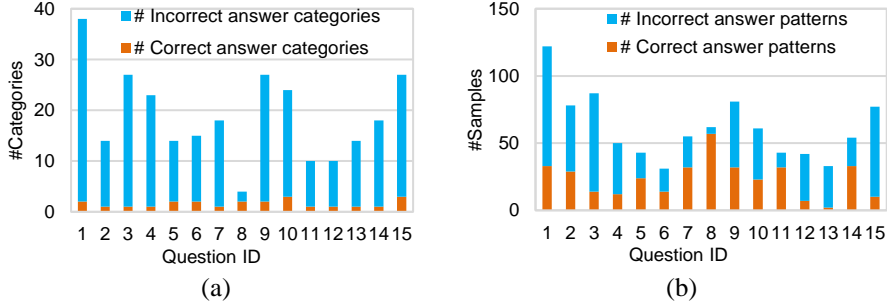

**Fig. 6.** Details of NIER_CBT. (a) shows the number of correct\incorrect categories in each question, and (b) shows the number of correct\incorrect patterns in each question.

## 5.2 OnHME recognizer

The overview of MTAP is shown in Fig. 7. It includes three main parts: the feature extraction, the encoder, and the decoder.

**Trajectory feature extraction.** We utilized the set of point-based features used in [15]. An OnHME is a sequence of trajectory points of pen-tip movements. We denote the sequence of $L$ points as $\{X_1, X_2, X_3, ..., X_L\}$ with $X_i = (x_i, x_i, s_i)$ where $(x_i, y_i)$ are the coordination of each point and $s_i$ is the corresponding stroke index of the i-th point. We store the sequence $\{X_i\}$ in the order of writing process. Before extracting the features, we firstly interpolate and normalize the original coordinates accoding to [20]. They are necessary to deal with non-uniform sampling in terms of writing speed and the size variations of the coordinate by using different devices to collect patterns. For each point, we extract an 8-dimensional feature vector as follows:

$$[x_i, y_i, dx_i, dy_i, d'x_i, d'y_i, \delta(s_i = s_{i+1}), \delta(s_i \neq s_{i+1})] \quad (8)$$

where $dx_i = x_{i+1} - x_i, dy_i = y_{i+1} - y_i, d'x_i = x_{i+2} - x_i, d'y_i = y_{i+2} - y_i$ and $\delta(\cdot) = 1$ when the conditional expression is true or otherwise zero, which presents the state of the pen (down/up).

**Encoder.** The encoder of MTAP is a combination of 4 stacked bidirectional GRUs [21] (BiGRUs) with a pooling operator, as shown in Fig. 7. Stacking multiple BiGRU layers could make the model learn high-level representation from the input. The input sequence of an upper BiGRU layer is the sequence of the hidden state of its lower BiGRU layer. Each layer has 250 forward and 250 backward units. Since the encoded features of two adjacent points are slightly different, the pooling layers are applied to reduce the complexity of the model and make the decoder part easier to parse with a fewer number of hidden states of the encoder. The pooling operator applied on the 2 top BiGRUs layers is to drop the even time steps of the lower BiGRU layer outputs and receive the odd outputs as the inputs. The hidden states outputted from the 4<sup>th</sup> layer are inputted to the decoder.

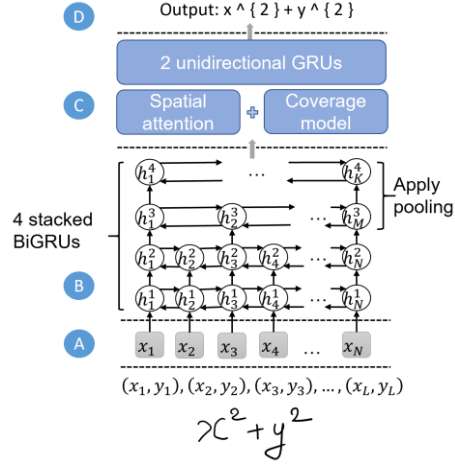

**Fig. 7.** Overview of TAP consisting of point-based features as the input (A), the encoder part (B), the decoder part (C), and the output (D).

**Decoder.** The MTAP decoder receives the hidden states $\{h_i\}_{i=\overline{1,K}}$ from the encoder and generates a corresponding LaTeX sequence of the input traces. The decoder consists of a word embedding layer of 256 dimensions, two layers of unidirectional GRU with 256 forward units, spatial attention, and a coverage model. The spatial attention points out the suitable local region in $\{h_i\}$ to attend for generating the next LaTeX symbol by assigning higher weights to a corresponding local annotation vector $\{a_i\}_{i=\overline{1,K}}$. The coverage model is a 1-dimensional convolutional layer to indicate whether a local region in $\{h_i\}$ has been attended to the generation. The model is trained with an attention guider using the oracle alignment information from the training OnHMEs to force the attention mechanism to learn well.

**Table 1.** ExpRate and CER of MTAP.

| Dataset | MTAP | | TAP [15] | |
|---|---|---|---|---|
| | **ExpRate (%)** | **CER (%)** | **ExpRate (%)** | **CER (%)** |
| CROHME 2014 testing set | 48.88 | 14.54 | 50.41 | 13.39 |
| Dset_Mix | 34.62 | 17.51 | - | - |
| NIER_CBT | 57.89 | 18.57 | - | - |

**Training and testing.** We trained MTAP by the training data of CROHME 2016. We removed genuine OnHME patterns in Dset_Mix for fair evaluation, because they are in the training data set. The optimizer and hyperparameters are the same as in [15]. Then, we measured the expression rate (ExpRate) and the character error rate (CER) on the testing data of CROHME 2014, Dset_Mix, and NIER_CBT, as shown in Table 1. The recognition rate of MTAP is 1.53 percentage points lower than the original TAP model on the CROHME 2014 testing set.

### 5.3 Experimental settings

We utilized the $k$-means algorithm and the complete linkage (CL) method for the clustering task. For $k$-means, we applied the Euclidean distance and initialized centroids using $k$-means++ [22], a popular variant of the $k$-means algorithm that tries to spread out initial centroids. To evaluate the proposed features, we set the number of clusters as the number of categories in our experiments.

### 5.4 Evaluation

In this section, we compare the proposed method with the previous methods. Moreover, we conduct experiments to evaluate our proposed SF.

**Comparison with other methods.** So far, clustering OnHMEs is not shown on the common dataset. We re-implemented the online bag-of-features proposed by Ung et al. [6] (denoted as *M1*) to compare with our proposed method. *M1* used an OnHME recognizer outputting a symbol relation tree instead of a LaTeX sequence, so that we used the OnHME recognizer proposed in [23] combined with an n-gram language model to extract the online bag-of-features. This recognizer achieves the expression rate of 51.70% on the CROHME 2014 testing set, which is 2.82 percentage points better than MTAP. In addition, we compared with the method using the edit distance, which is to compute the dissimilarity between two LaTeX sequences outputted from MTAP, denoted as *M2*. SbRs produced by the edit distance are inputted to the $k$-means algorithm.

We carried out several experiments to evaluate the effectiveness of GSSF and SbR on the representation. Firstly, we directly used the absolute of GSSF as the distance function to input into CL, denoted as *M3*. Secondly, we used the SbR matrix produced by MTAP and GSSF to input into CL by using the Euclidean distance, denoted as *M4*.

Thirdly, we used the SbR matrix produced by MTAP and GSSF to input into the *k*-means algorithm, denoted as *M5*.

We also compared our proposed method with previous methods for clustering Of-HMEs, which consists of the offline bag-of-features proposed by Khuong et al. [5] (denoted as *M6*) and the CNN-based features proposed by Nguyen et al. [7] (denoted as *M7*). OfHMEs are converted from OnHMEs. We used a symbol classifier to extract the offline bag-of-features in *M6*. We also trained a CNN model to extract spatial classification features for *M7*. Those models in *M1*, *M6*, and *M7* were trained in the same dataset with MTAP.

**Table 2.** Comparisons with other methods of clustering HMEs. Values are presented in form of "average value (standard deviation)"

| HME type | Name | Features | Clustering algorithm | Dset_Mix | | NIER_CBT | |
|---|---|---|---|---|---|---|---|
| | | | | **Purity** | *MC* | **Purity** | *MC* |
| OnHME | Ung et al. (*M1*) | Online bag-of-features | *k*-means | 0.764 (0.11) | 0.629 (0.09) | 0.895 (0.05) | 0.712 (0.06) |
| | Ours (*M2*) | MTAP + Edit distance + SbR | *k*-means | 0.638 (0.12) | 0.700 (0.05) | 0.867 (0.05) | 0.725 (0.07) |
| | Ours (*M3*) | MTAP + GSSF | CL (GSSF) | 0.806 (0.10) | 0.611 (0.05) | **0.921** **(0.05)** | 0.698 (0.06) |
| | Ours (*M4*) | MTAP + GSSF + SbR | CL (Euclidean) | 0.775 (0.15) | 0.633 (0.07) | **0.920** **(0.04)** | 0.700 (0.06) |
| | **Ours (*M5*)** | **MTAP + GSSF + SbR** | *k*-means | **0.916** **(0.05)** | **0.556** **(0.03)** | 0.915 (0.03) | **0.702** **(0.07)** |
| OfHME | Khuong et al. (*M6*) | Offline bag-of-features | *k*-means | 0.841 (0.15) | 0.595 (0.07) | 0.834 (0.07) | 0.739 (0.08) |
| | Nguyen et al. (*M7*) | CNN-based features | *k*-means | 0.723 (0.16) | 0.653 (0.08) | 0.829 (0.06) | 0.744 (0.06) |

Table 2 shows that our proposed GSSF combined with MTAP, i.e., *M3*, *M4*, and *M5*, outperforms *M1*, *M2*, and *M7* in purity and *MC* on both Dset_Mix and NIER_CBT. *M3* and *M4* have lower performance than *M6* on Dset_Mix. *M5* yields the best performance on Dset_Mix, while *M3* performs best on NIER_CBT. However, *M5* achieves a high purity on both datasets. Regarding *MC*, *M5* achieves the marking cost of around 0.556 and 0.702 in Dset_Mix and NIER_CBT. Consequently, the marking cost is reduced by 0.444 and 0.298 than manual marking.

There are some discussions based on the results of *M3*, *M4*, and *M5*. Firstly, our GSSF without SbR works well when using the CL method on NIER_CBT. Secondly, GSSF combined with SbR achieves more stable performance when using the *k*-means algorithm than the CL method with the Euclidean distance.

**Evaluations on our similarity function.** We conducted experiments on forming our proposed GSSF. According to Eq. (3), GSSF is formed by taking an average of two similarity components $F(S_1|S_2)$ and $F(S_2|S_1)$ since we aim to make SF symmetric. We compare GSSF with three possible variants as follows:

- We directly use function $F(x|y)$ as SF so that SbR of $X_i$ in Eq. (4) is as $[F(X_i|X_1), \dots, F(X_i|X_i), \dots, F(X_i|X_N)]$. Since $F$ is not a symmetric function, we denote this SF as Asymmetric_GSSF.

- We define two SFs by getting the minimum and maximum value between $F(S_1|S_2)$ and $F(S_2|S_1)$ instead of taking the average of them. We denote them as Min_GSSF and Max_GSSF, respectively.

**Table 3.** Comparisons with other variants of SFs

| Method | Purity | |
|---|---|---|
| | Dset_Mix | NIER_CBT |
| Asymmetric_GSSF | 0.857±0.07 | 0.907±0.05 |
| Min_GSSF | 0.861±0.08 | 0.907±0.04 |
| Max_GSSF | 0.909±0.08 | 0.913±0.04 |
| **GSSF** | **0.916±0.05** | **0.915±0.03** |

Table 3 presents the performance of our proposed SF with Asymmetric_GSSF, Min_GSSF, and Max_GSSF in terms of purity by using the $k$-means algorithm. Our GSSF performs better than the other SFs on Dset_Mix and NIER_CBT, which implies that taking the average of two similarity components is better than using them directly or taking the minimum or maximum value between them. However, Max_GSSF yields comparable results with GSSF.

### 5.5 Visualizing similarity-based representation matrix

This section shows the visualization of the SbR matrix to see how this representation discriminates for clustering. Fig. 8 presents the SbR matrix of the subgroup 3 and 8 in Dset_Mix. OnHMEs belonging to the same class are placed together in both dimensions. For subgroup 3, its categories are significantly distinct. We can see that SbR well represents for OnHMEs in the same category. The similarity scores among intra-category OnHMEs almost near 0, and they are much higher than those among inter-category OnHMEs. On the other hand, some categories in subgroup 8 are slightly different, such as categories $C_3$ and $C_5$ or category $C_4$ and $C_6$, and SbR among them is not so different. Purity on subgroups 3 and 8 are 0.984 and 0.832, respectively.

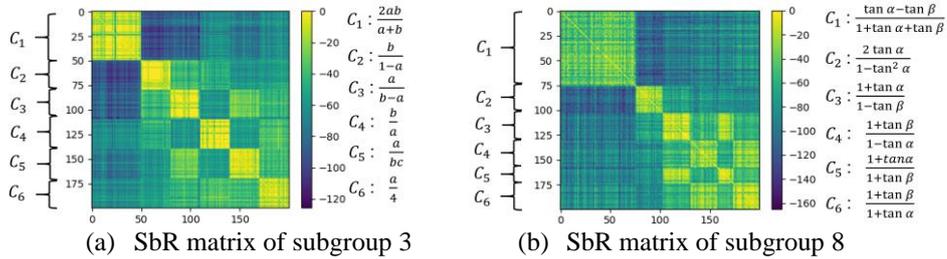

(a) SbR matrix of subgroup 3    (b) SbR matrix of subgroup 8

**Fig. 8.** Visualization of the SbR matrix of the subgroup 3 and 8 before normalizing them into $[0, 1]$.

# 6 Conclusion and future works

This paper presented a similarity-based representation (SbR) and a generative sequence similarity function (GSSF) based on the Seq2Seq recognizer for clustering to provide computer-assisted marking for handwritten mathematics answers OnHMEs. The SbR matrix is then inputted to the $k$-means algorithm by setting the number of clusters as the number of categories. We achieved around 0.916 and 0.915 for purity and around 0.556 and 0.702 for the marking cost on the two answer datasets, Dset_Mix and NIER, respectively. Our method outperforms other methods on clustering HMEs.

There are several remaining works. We need to study the method to estimate the correct number of clusters and consider mini-batch clustering for larger answer sets. User interface for markers is also another problem to study.

## Acknowledgement

This research is being partially supported by the grant-in-aid for scientific research (A) 19H01117 and that for Early Career Research 18K18068.

## References

1. M. Mahdavi, R. Zanibbi, H. Mouchere, C. Viard-Gaudin, U. Garain: CROHME + TFD: Competition on recognition of handwritten mathematical expressions and typeset formula detection. In: Proc. of Inter. Conf. on Doc. Anal. and Recognit. pp. 1533–1538 (2019).
2. LaViola, J.J., Zeleznik, R.C.: MathPad2: A system for the creation and exploration of mathematical sketches. ACM Trans. Graph. 23, 432–440 (2004).
3. K. F. Chan, D. Y. Yeung: PenCalc: a novel application of on-line mathematical expression recognition technology. In: Proc. of International Conference on Document Analysis and Recognition. pp. 774–778 (2001).
4. T. O'Connell, C. Li, T. S Miller., R. C. Zeleznik, J. J. LaViola: A usability evaluation of AlgoSketch: a pen-based application for mathematics. In: Proc. of Eurographics Symp. Sketch-Based Interfaces Model. pp. 149–157 (2009).
5. V. T. M. Khuong, K. M. Phan, H. Q. Ung, C. T. Nguyen, M. Nakagawa: Clustering of Handwritten Mathematical Expressions for Computer-Assisted Marking. IEICE Trans. Inf. Syst. E104.D, 275–284 (2021). https://doi.org/10.1587/transinf.2020EDP7087.
6. H. Q. Ung, V. T. M. Khuong, A. D. Le, C. T. Nguyen, M. Nakagawa: Bag-of-features for clustering online handwritten mathematical expressions. In: Proc. of Inter. Conf. Pattern Recognit. Artif. Intell. pp. 127–132 (2018).
7. C. T. Nguyen, V. T. M. Khuong, H. T. Nguyen, M. Nakagawa: CNN based spatial classification features for clustering offline handwritten mathematical expressions. Pattern Recognit. Lett. 131, 113–120 (2020).
8. D. François, V. Wertz, M. Verieysen: The concentration of fractional distances. IEEE Trans. Knowl. Data Eng. 19, 873–886 (2007).

9.  R. Cummins, M. Zhang, T. Briscoe: Constrained Multi-Task Learning for Automated Essay Scoring. In: Proc. of Annu. Meet. Assoc. Comput. Linguist. pp. 789–799 (2016).

10. V. Salvatore, N. Francesca, C. Alessandro: An Overview of Current Research on Automated Essay Grading. J. Inf. Technol. Educ. Res. 2, 319–330 (2003).

11. T. Ishioka, M. Kameda: Automated Japanese essay scoring system:jess. In: Proc. of Int. Work. Database Expert Syst. Appl. pp. 4–8 (2004).

12. S. Srihari, J. Collins, R. Srihari, H. Srinivasan, S. Shetty, J. Brutt-Griffler: Automatic scoring of short handwritten essays in reading comprehension tests. Artif. Intell. 172, 300–324 (2008).

13. S. Basu, C. Jacobs, L. Vanderwende: Powergrading: a Clustering Approach to Amplify Human Effort for Short Answer Grading. Trans. Assoc. Comput. Linguist. 1, 391–402 (2013).

14. M. Brooks, S. Basu, Jacobs, C., Vanderwende, L.: Divide and correct: Using clusters to grade short answers at scale. In: Proc. of ACM conf. on Learning @ Scale. pp. 89–98 (2014).

15. J. Zhang, J. Du, L. Dai: Track, Attend, and Parse (TAP): An End-to-End Framework for Online Handwritten Mathematical Expression Recognition. IEEE Trans. Multimed. 21, 221–233 (2019).

16. Z. Hong, N. You, J. Tan, N. Bi: Residual BiRNN based Seq2Seq model with transition probability matrix for online handwritten mathematical expression recognition. In: Proceedings of the International Conference on Document Analysis and Recognition, ICDAR. pp. 635–640 (2019). https://doi.org/10.1109/ICDAR.2019.00107.

17. V. T. M. Khuong: A Synthetic Dataset for Clustering Handwritten Math Expression TUAT (Dset_Mix) - TC-11, http://tc11.cvc.uab.es/datasets/Dset_Mix_1.

18. K. M. Phan, V. T. M. Khuong, H. Q. Ung, M. Nakagawa: Generating Synthetic Handwritten Mathematical Expressions from a LaTeX Sequence or a MathML Script. In: Proc. of International Conference on Document Analysis and Recognition. pp. 922–927 (2020).

19. F. Yasuno, K. Nishimura, S. Negami, Y. Namikawa: Development of Mathematics Items with Dynamic Objects for Computer-Based Testing Using Tablet PC. Int. J. Technol. Math. Educ. 26, 131–137 (2019).

20. X. Y. Zhang, F. Yin, F. Y. Zhang, C. L. Liu, Y. Bengio: Drawing and Recognizing Chinese Characters with Recurrent Neural Network. IEEE Trans. Pattern Anal. Mach. Intell. 40, 849–862 (2018).

21. K. Cho, B. V. Merriënboer, C. Gulcehre, D. Bahdanau, F. Bougares, H. Schwenk, Y. Bengio: Learning phrase representations using RNN encoder-decoder for statistical machine translation. In: Conference on Empirical Methods in Natural Language Processing 2014. pp. 1724–1734 (2014). https://doi.org/10.3115/v1/d14-1179.

22. D. Arthur, S. Vassilvitskii: k-means++: the advantages of careful seeding. In: Proc. of Annu. ACM-SIAM Symp. Discret. algorithms. pp. 1027–1035 (2007).

23. C. T. Nguyen, T. N. Truong, H. Q. Ung, M. Nakagawa: Online Handwritten Mathematical Symbol Segmentation and Recognition with Bidirectional Context. In: Proc. of Int. Conf. Front. Handwrit. Recognit. pp. 355–360 (2020).

24. Lenco, D., Interdonato, R.: Deep Multivariate Time Series Embedding Clustering via

Attentive Gated Autoencoder. In: Lecture Notes in Computer Science. pp. 318–329 (2020).

25. Ma, Q., Zheng, J., Li, S., Cottrell, G.W.: Learning Representations for Time Series Clustering. In: Advances in Neural Information Processing Systems. pp. 3776–3786 (2019).

26. S. J. Rao, Y. Wang, G. Cottrell: A Deep Siamese Neural Network Learns the Human-Perceived Similarity Structure of Facial Expressions Without Explicit Categories. In: Proceedings of the 38th Annual Conference of the Cognitive Science Society. pp. 217–222 (2016).